\documentclass{article}

\PassOptionsToPackage{numbers}{natbib}
\usepackage[final,main]{accord}

\usepackage[utf8]{inputenc}
\usepackage[T1]{fontenc}
\usepackage{hyperref}
\usepackage{url}
\usepackage{booktabs}
\usepackage{amsfonts}
\usepackage{nicefrac}
\usepackage{microtype}
\usepackage{xcolor}
\usepackage{graphicx}
\usepackage{subcaption}
\usepackage[most]{tcolorbox}
\usepackage{tikz}
\usetikzlibrary{arrows.meta, positioning, shapes.geometric, calc, fit, backgrounds}
\definecolor{rqBlueBg}{RGB}{235,243,255}
\definecolor{rqBlueDark}{RGB}{15,35,90}
\tcbset{
  aibox/.style={
    width=\linewidth,
    top=7pt,
    bottom=2pt,
    colback=rqBlueBg,
    colframe=rqBlueDark,
    colbacktitle=rqBlueDark,
    enhanced,
    center,
    attach boxed title to top left={yshift=-0.1in,xshift=0.15in},
    boxed title style={boxrule=0pt,colframe=white,},
  }
}
\newcounter{takeaway}
\newtcolorbox{takeaway}[1][]{
  aibox,
  coltitle=white,
  fonttitle=\bfseries,
  title={\stepcounter{takeaway}Takeaway \thetakeaway},
  #1
}
\usepackage{algorithm}
\usepackage{algorithmic}
\usepackage{placeins}
\usepackage{pdfpages}
\usepackage{fancyvrb}
\usepackage{fvextra}

\usepackage{titlesec}
\titlespacing*{\section}{0pt}{1.4ex plus 0.6ex minus 0.2ex}{0.7ex plus 0.2ex}
\titlespacing*{\subsection}{0pt}{1.0ex plus 0.5ex minus 0.2ex}{0.5ex plus 0.2ex}
\titlespacing*{\subsubsection}{0pt}{0.8ex plus 0.4ex minus 0.2ex}{0.4ex plus 0.2ex}
\titlespacing*{\paragraph}{0pt}{0.6ex plus 0.3ex minus 0.1ex}{0.6em}

\newcommand{\accord}{\textsc{Accord}}
\newcommand{\term}[1]{\emph{#1}}

\definecolor{orange}{rgb}{1,0.5,0}
\definecolor{mdgreen}{rgb}{0.05,0.6,0.05}
\definecolor{mdblue}{rgb}{0,0,0.7}
\definecolor{dkblue}{rgb}{0,0,0.5}
\definecolor{dkgray}{rgb}{0.3,0.3,0.3}
\definecolor{slate}{rgb}{0.25,0.25,0.4}
\definecolor{gray}{rgb}{0.5,0.5,0.5}
\definecolor{ltgray}{rgb}{0.7,0.7,0.7}
\definecolor{purple}{rgb}{0.7,0,1.0}
\definecolor{lavender}{rgb}{0.65,0.55,1.0}

\definecolor{mypurple}{RGB}{111,61,121}
\definecolor{myblue}{RGB}{46,88,180}
\definecolor{myred}{RGB}{181,68,106}
\definecolor{myyellow}{RGB}{204,143,55}

\newif\ifshowcomments
\showcommentsfalse

\definecolor{lailightblue}{RGB}{86,156,214}
\NewDocumentCommand{\lai}
{ mO{} }{\ifshowcomments\textcolor{lailightblue}{\textsuperscript{\textit{Lai}}\textsf{\textbf{\small[#1]}}}\fi}
\NewDocumentCommand{\ember}
{ mO{} }{\ifshowcomments\textcolor{orange}{\textsuperscript{\textit{Ember}}\textsf{\textbf{\small[#1]}}}\fi}
\NewDocumentCommand{\cheng}
{ mO{} }{\ifshowcomments\textcolor{orange}{\textsuperscript{\textit{Cheng}}\textsf{\textbf{\small[#1]}}}\fi}

\NewDocumentCommand{\heng}
{ mO{} }{\ifshowcomments\textcolor{red}{\textsuperscript{\textit{Heng}}\textsf{\textbf{\small[#1]}}}\fi}

\title{\textsc{Accord}: Action-Conditioned Contextual Grounding for Language Agents}

\author{%
  Lai Jiang\textsuperscript{1} \quad
  Cheng Qian\textsuperscript{1} \quad
  Zhenhailong Wang\textsuperscript{1} \quad
  Pan Lu\textsuperscript{2} \quad
  Heng Ji\textsuperscript{1} \quad
  Hao Peng\textsuperscript{1} \\
  \textsuperscript{1}University of Illinois Urbana-Champaign \\
  \textsuperscript{2}Stanford University
}

\begin{document}

\maketitle
\begingroup
\renewcommand\thefootnote{}
\footnotetext{Project page: \url{https://github.com/Jianglai-0023/ACCORD}}
\endgroup

\begin{abstract}
User instructions are often underspecified because humans rely on implicit assumptions about the surrounding environment. For large language model (LLM) agents operating in information-rich digital and physical environments, these assumptions cannot be inferred from the instruction alone; they must be recovered from the current state of tools, data, interfaces, and observations. Effective execution therefore requires agents to identify missing context, ground it in observed evidence, and carry it forward into subsequent actions.
We show that current agents often fail to do so. They act from assumed rather than observed specifics, overlook information they could have gathered, and fail to incorporate evidence that has already been returned.
Building on this insight, we propose \accord{} (\textbf{A}ction-\textbf{C}onditioned \textbf{Co}ntextual G\textbf{r}oun\textbf{d}ing), a simple and effective agent framework for \emph{adaptive grounding}. 
Before each action, \accord{} actively probes the environment for missing information and integrates relevant context from the agent's trajectory that would otherwise be overlooked.
Requiring no additional training or task-success signals, \accord{} improves task-goal completion on AppWorld by up to $+20.6$ points with GPT-5-mini, from $42.0\%$ to $62.6\%$, compared to strong baselines. These gains persist with a substantially stronger base model ($+10.8$ with Claude-4.5-sonnet), an open-weight model ($+10.1$ with Qwen3.5-27B-FP8), and on the embodied AlfWorld benchmark ($+7.4$ success rate with GPT-5-mini).
\end{abstract}

\section{Introduction}
\label{sec:intro}


\begin{figure}[t]
    \centering
    \includegraphics[width=\textwidth]{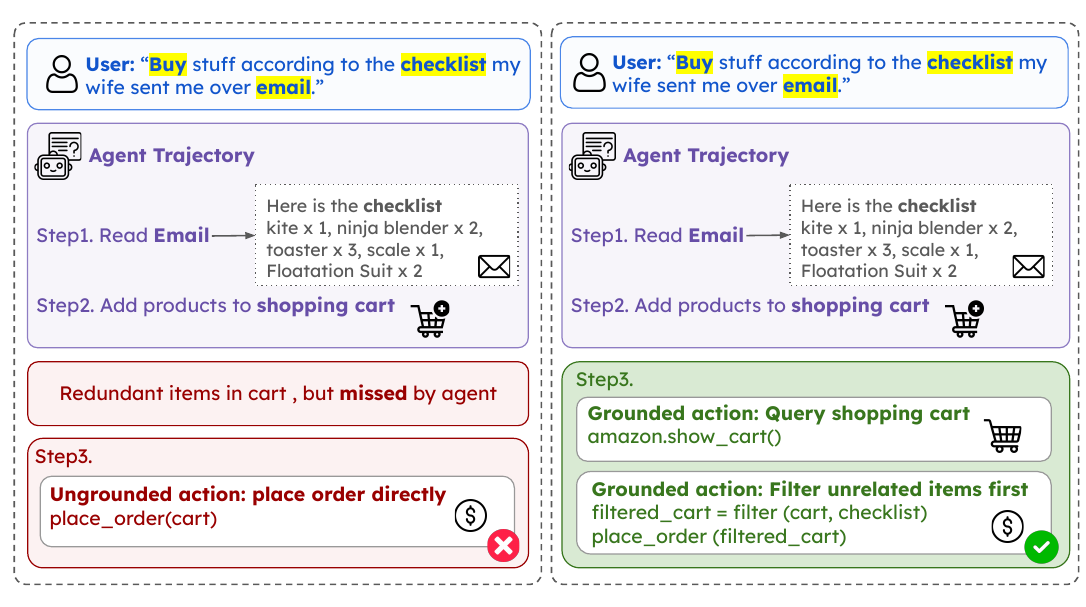}
    \caption{Ungrounded agent behavior (\textbf{left}) vs.\
    grounded behavior in the environment (\textbf{right}). The user gives an underspecified instruction whose intended
    meaning is fixed by the environmental state. Yellow highlights indicate the objects that need further clarification.
    Both panels share the same prior trajectory. Crucially, the
    cart also contains unrelated items, leaving the agent's
    trajectory \emph{incomplete} at decision time.
    \textbf{(Ungrounded)} The agent issues \texttt{place\_order(cart)} directly; the redundant
    items are submitted unnoticed.
    \textbf{(Grounded)} The agent first probe
    \texttt{amazon.show\_cart()} to surface the cart state, then
    \texttt{filter(cart, checklist)} and
    \texttt{place\_order(filtered\_cart)} so only the matching items
    are submitted.}
    \label{fig:teaser}
\end{figure}


Large language models (LLMs) are increasingly deployed as agents that
translate a user's natural-language request into sequences of tool and
API calls, ranging from coding assistants~\citep{yang2024sweagent,jimenez2024swebench}
to web and personal-app agents~\citep{deng2023mind2web,zhou2024webarena,trivedi2024appworld},
all built on top of general-purpose tool-use mechanisms~\citep{schick2023toolformer,patil2023gorilla,qin2024toolllm}.
The success of every such deployment rests on the agent correctly
inferring what they should do from a short natural-language prompt and the environment.

Yet communication in natural language is inherently context dependent:
humans converse against a backdrop of implicit assumptions and shared
knowledge about the surrounding environment, and rarely spell out the
concrete entities, constraints, and procedural details needed to act on
a request~\citep{shaikh2024groundinggaps,tamkin2023ambiguity}.
User prompts thus often arrive \emph{underspecified}, admitting
multiple plausible interpretations of intent.
As a result, without grounding in the environment, the agent overlooks
the disambiguating context and often acts on the wrong interpretation.

Consider the shopping-assistant task in Figure~\ref{fig:teaser}: the
user asks the agent to order the items on an emailed checklist, but the
cart already contains three off-checklist items from earlier browsing.
The prompt alone does not tell which set of items the user means; the
resolution lives in the environment itself, in the \texttt{cart} state
and the \texttt{filter} affordance, not in the instruction.

We argue that as LLMs are embedded in increasingly rich digital and
physical environments~\citep{trivedi2024appworld,yao2025taubench,zhou2024webarena},
every \emph{write} action that commits state to the environment should be explicitly \emph{grounded} in
the contextual signals surfaced by ongoing interaction with the
environment.
Executing an instruction increasingly requires \emph{action
grounding}: at every step, the agent must align the next \emph{write}
with the environmental state that resolves the underspecified parts
of the request.

Current LLM agent frameworks treat the environment in two predominant
modes: as a site of task execution, where actions are issued and tools
are invoked~\citep{yao2023react,schick2023toolformer,patil2023gorilla,qin2024toolllm},
and as a source of feedback for self-correction or wider
search~\citep{shinn2023reflexion,gou2024critic,zhou2024lats,hao2023rap}.
What remains under-explored is treating the environment as a \emph{source
of information for resolving the user's underspecified intent}.

Two structural gaps keep this
kind of grounding from happening reliably, and they map directly onto
two failure modes that we show that are common among current agent systems: Figure~\ref{fig:error-distribution};\S\ref{sec:gap-missing}--\ref{sec:gap-overlooked}.
\emph{On the input side}, agents fall back on priors likely shaped by
training on similar instructions, hallucinating tool affordances and
argument values rather than probing the environment, so the
trajectory ends up \emph{incomplete}, lacking the information the
upcoming write requires.
\emph{On the feedback side}, the environment returns observations
formatted for its own purposes, not for the instruction: schemas are
verbose, decisive facts are buried in long
responses~\citep{liu2024lostinmiddle,du2025contextlength}, and key
signals are conveyed only implicitly (e.g., a success code without the
resulting state), leaving the information \emph{present but overlooked} at the decision point.
We refer to these two modes as
\emph{incomplete} and \emph{overlooked} grounding failures, and target
both with a unified inference-time mechanism (\S\ref{sec:method}).

Building on this analysis, we adopt a simple but effective approach
that closes both grounding gaps at two complementary layers of the
agent system (\S\ref{sec:method}). At the inference-time layer
(\S\ref{sec:grounding-agent}), our framework resurfaces existing
trajectory evidence and probes the environment facts, so that the
agent's context contains the information needed for the upcoming
action. At the policy layer (\S\ref{sec:grounding-prompt}), we
additionally encourage the main agent's to actively probe and explore
the environment, so that its trajectory carries enough environmental
information for the inference-time layer to draw on. Together, the
two layers ensure the agent acts on the right interpretation of the
user's intent. We call this two-layer framework \accord{}
(\textbf{A}ction-\textbf{C}onditioned \textbf{Co}ntextual G\textbf{r}oun\textbf{d}ing).

At the inference-time layer, \accord{} intervenes before any
\emph{write} action being executed by the environment, pausing the agent and constructing
an \emph{action-ready grounded state}: a context include information required by the
upcoming action. Crucially, \accord{} grounds writes in \emph{objective signals from the
environment itself}, such as live observations, returned data, and
tool schemas, rather than in the agent's own subjective self-critique,
as in reflection- or refine-style
approaches~\citep{shinn2023reflexion,gou2024critic} that iterate over
the model's internal beliefs. In unseen environments, the agent's
internal model is unreliable, so reflecting on that
model cannot resolve the underlying information deficit. Moreover, \accord{}'s policy layer modifies the
main agent's policy with a prompt that encourages it to gather
sufficient environmental information early in the trajectory. 
The two layers play distinct roles. The inference-time layer is
where grounding actually happens and the policy layer is supporting. Both layers are training-free and model-agnostic,
allowing the same mechanism to transfer unchanged across tasks,
environments, and underlying base models.

We evaluate \accord{} on AppWorld and ALFWorld across three LLMs:
closed-weight GPT-5-mini and Claude-4.5-sonnet, and open-weight
Qwen3.5-27B-FP8. Across all settings, \accord{} improves
task completion over a ReAct baseline by $7.4\%$--$20.6\%$, with the
largest gains on the harder AppWorld test-challenge split: $+20.6$ TGC
on GPT-5-mini and $+16.1$ on Qwen3.5-27B-FP8. \accord{}
generalizes to ALFWorld, improving Qwen3.5-27B-FP8 from $80.7\%$
to $88.1\%$ success rate.

Our work reveals that whether an agent can effectively probe and
use environmental information is a decisive factor in its
performance. 
We hope this work helps push forward the understanding of the
meta-capabilities agents will need to develop, and prompts further
thinking about how agents
can better interact with their environments and manage their own
context, so that they can operate effectively in continuously
changing environments with minimal human intervention.

\section{Problem Formulation and Empirical Analysis}

\subsection{Action selection in interactive environments}

We consider an LM agent operating in an interactive environment $\mathcal{E}$
to accomplish a natural-language task $\tau$.
At each step $t$, the agent selects an action $a_t \in \mathcal{A}$ based on
the context $c_t = (\tau, a_0, o_0, \ldots, a_{t-1}, o_{t-1})$,
where $o_i$ is the observation returned by $\mathcal{E}$ after $a_{i}$.
The action space $\mathcal{A}$ varies by environment: in our experiments,
an action is a Python snippet that invokes one or more APIs (AppWorld)
or a natural-language command such as ``pick up X'' (ALFWorld).
As in real-world deployment, the agent receives no task-level reward or
success signal during execution: all feedback comes through the
environment's own observations $o_i$, which may include API error
We single out the subset of \term{write} actions
$\mathcal{W} \subseteq \mathcal{A}$ that alter the environment state:
these are typically irreversible and largely determine task success (e.g., place order, delete emails).

\paragraph{Action grounding.}
The central question we study is whether each write $w$ is properly
grounded in the information available at the time of the
decision. We say $c_t$ is action-ready for a candidate write
$w$, denoted $\mathrm{ActionReady}(c_t, w)$, when $c_t$
\emph{exposes} the affordance, argument values, and constraints
relevant to $w$, \emph{and} $w$ is \emph{consistent with} those
signals (no element of $w$ contradicts what $c_t$ exposes).
Otherwise we say there is a grounding gap between $c_t$
and $w$, arising in two distinct ways: incomplete, when
the required information has not entered $c_t$, and
overlooked, when $w$ contradicts information already in
$c_t$. The next two subsections characterize each mode in turn,
quantifying its prevalence on ALFWorld and AppWorld and identifying
recurring failure patterns.

\begin{figure}[t]
    \centering
    \begin{minipage}[t]{0.48\textwidth}
        \centering
        \includegraphics[width=\linewidth]{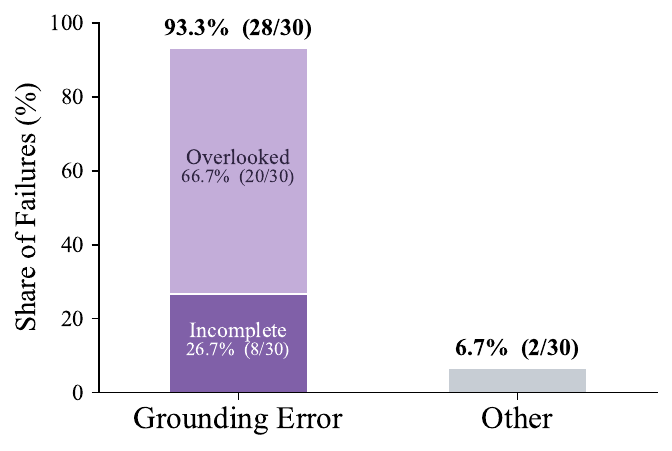}
        \subcaption{ALFWorld ($n=30$).}
        \label{fig:error-distribution-alfworld}
    \end{minipage}\hfill
    \begin{minipage}[t]{0.48\textwidth}
        \centering
        \includegraphics[width=\linewidth]{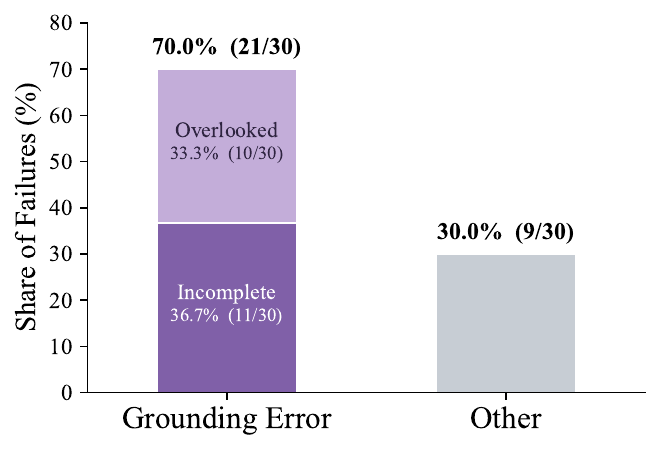}
        \subcaption{AppWorld ($n=30$).}
        \label{fig:error-distribution-appworld}
    \end{minipage}
    \caption{Distribution of grounding-gap types among baseline ReAct (GPT-5-mini) failures on ALFWorld~(\subref{fig:error-distribution-alfworld}) and AppWorld~(\subref{fig:error-distribution-appworld}). \emph{Overlooked} (Section~\ref{sec:gap-overlooked}): the required information was already present in $c_t$ but not used. \emph{Incomplete} (Section~\ref{sec:gap-missing}): the information was never acquired before the write. \emph{Other}: residual failures not attributable to a grounding gap.
    }
    \label{fig:error-distribution}
\end{figure}

\subsection{Incomplete: Writes ungrounded in the environment}
\label{sec:gap-missing}

In \textbf{incomplete} failures, $c_t$ does not yet carry the
affordance, arguments, or constraints needed to execute $w$ correctly,
and the agent writes anyway rather than first probing $\mathcal{E}$ for
the missing facts.
The required information has not yet entered the trajectory, so
any $w$ the agent issues against this under-specified picture of
the environment is effectively a hallucination rather than a
grounded decision.

\paragraph{Quantifying the gap.}
AppWorld~\citep{trivedi2024appworld} is an API-based LLM-agent
benchmark spanning nine interconnected everyday applications, and
ALFWorld~\citep{shridhar2021alfworld} is a text-based embodied
benchmark of household tasks.
We randomly sample 30 failed trajectories from the GPT-5-mini ReAct
baseline on each, and annotate each failed write $w$ for whether
the information required to execute $w$ correctly was already
present in $c_t$ at the time $w$ was issued. AppWorld annotations
are produced by Claude Opus~4.6 with a multi-label rubric;
ALFWorld annotations are produced by a rule-based string-matching
script. Both rubrics and per-class counts are provided in
Appendix~\ref{app:annotation}. A consistent quarter of failures
occur under an \textbf{incomplete} gap, where the agent had not
acquired the required information before acting:
26.7\% (8/30) on ALFWorld and 36.7\% (11/30) on AppWorld
(Figure~\ref{fig:error-distribution}, ``Incomplete'').

\paragraph{Why agents act on incomplete information.}
We identify two recurring patterns from the annotated failures,
both instances where the agent never probed for the required
information.
First, agents \emph{assume} environmental content rather than reading it:
having located a relevant resource (e.g., a note, a document, a record),
the agent skips inspecting its content and writes a parser or downstream
action against an assumed format.
Second, agents \emph{stop at known affordances}: having handled
the task through APIs they have already seen, the agent does not
probe $\mathcal{E}$ for additional APIs the task may also require,
leaving parts of the task undone. For example, asked to delete a
payment card across applications, the agent deletes it in one app
and assumes no other app stores one, never inspecting the other
apps' APIs.
Both patterns reflect the same underlying behavior---the agent treats
information acquisition as optional when its prior is confident enough,
even when the prior is uncalibrated to the current environment.

\subsection{Overlooked: Writes ungrounded in the prior trajectory}
\label{sec:gap-overlooked}

In \textbf{overlooked} failures, the required information has
\emph{already} entered $c_t$ through earlier observations, but the
agent issues a $w$ whose content contradicts that information. No additional probe is needed; the
context already carries what is needed, but the agent writes as if it
did not.

\paragraph{Quantifying the gap.}
Annotations follow \S\ref{sec:gap-missing} (full rubrics and
per-class counts in Appendix~\ref{app:annotation}). The overlooked case accounts for a
substantial fraction of failures on both benchmarks: 66.7\%
(20/30) on ALFWorld and 33.3\% (10/30) on AppWorld
(Figure~\ref{fig:error-distribution}, ``Overlooked''). This
indicates that simply acquiring more information is insufficient:
the agent must also be steered to use what it has already
observed.


\paragraph{Why agents overlook what they already have.}
Two conditions make overlooking more likely, and together they suggest why
acquired information fails to ground writes.
First, \emph{distance}: the information relevant to the current
write often lives many steps away from the write itself.
This echoes recent findings that LM performance on long-context
tasks degrades with input length even when the relevant content is
already present in the
input~\citep{liu2024lostinmiddle,levy2024sametask,hsieh2024ruler,du2025contextlength}.
Second, \emph{format mismatch}: information returned in a form that diverges
from the write's argument schema---buried in a long, noisy response with
many irrelevant fields, or conveyed only implicitly. This
echoes work showing that LLM reasoning degrades when relevant facts are
embedded in irrelevant or noisy
context~\citep{shi2023distracted,wu2024irrelevant,cuconasu2024noise}.
Both observations point to the conclusion that whether $c_t$ is
action-ready depends not only on what it contains but on whether the
contents are in a form the agent can attend to.
This motivates treating overlooked grounding as a problem of
\emph{re-surfacing} relevant prior information at decision time, which our
method addresses directly.

\paragraph{Summary.}
Both subsections describe a write action
issued without sufficient grounding.
Our method targets both forms under a single mechanism for closing the
grounding gap before each write.

\FloatBarrier
\section{\accord{}: Action-Conditioned Contextual Grounding}
\label{sec:method}

\begin{figure}[t]
\centering
\includegraphics[width=\textwidth]{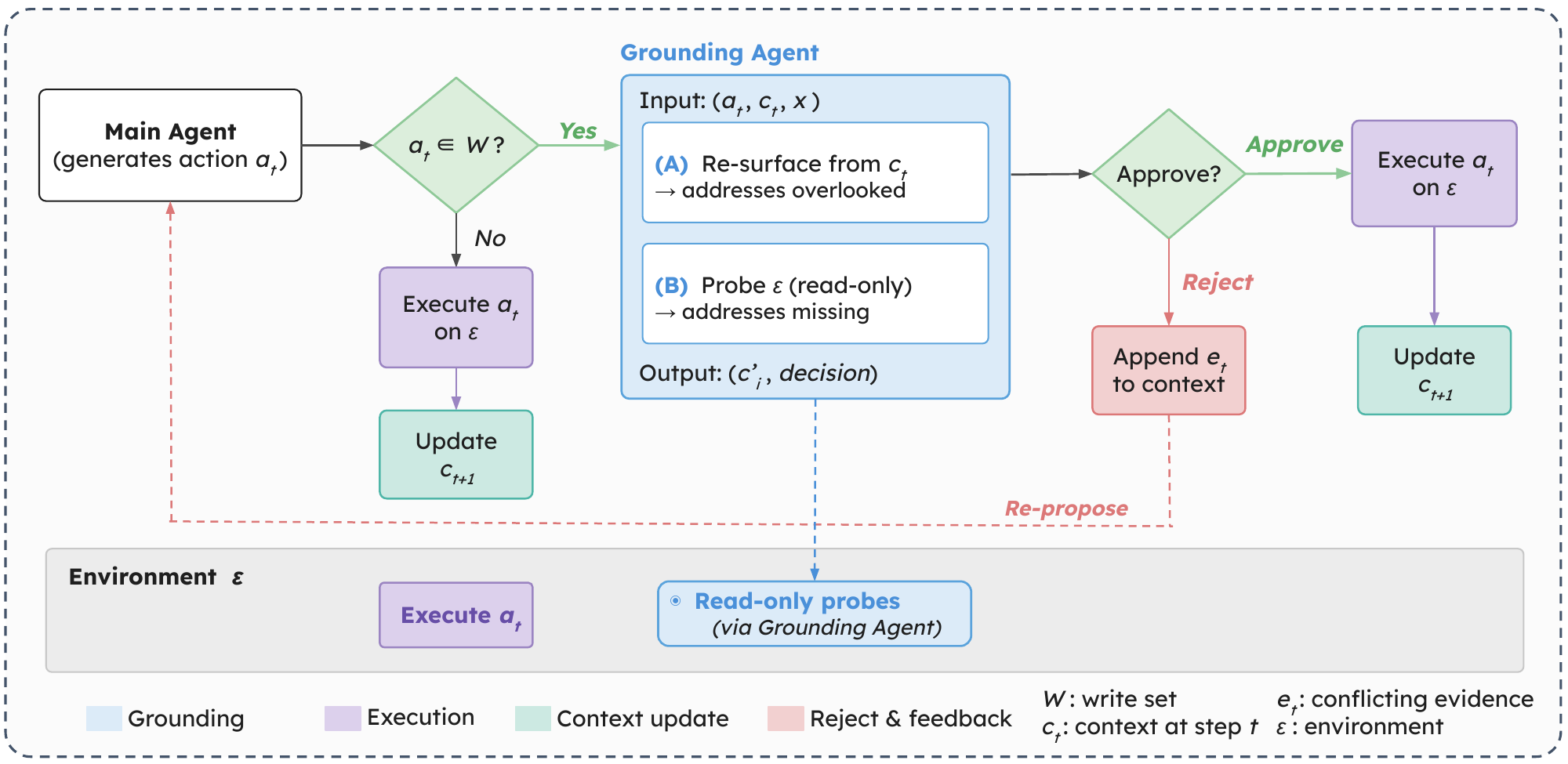}
\caption{\accord{}'s adaptive grounding loop (full pseudocode in
Algorithm~\ref{alg:framework}). At each
step the main agent proposes a write $w_t$; if $w_t \in \mathcal{W}$
(a write action), the grounding agent (blue) receives
$(w_t, c_t, x)$ and augments the context via two complementary moves:
\emph{(A)} re-surfacing trajectory evidence (addressing the
\textit{overlooked} mode) and \emph{(B)} probing $\mathcal{E}$ via
\emph{read-only} calls (addressing the \textit{incomplete} mode). On
\textsc{approve}, $w_t$ executes on $\mathcal{E}$; on \textsc{reject}, the conflicting evidence $e_t$ is
appended and the main agent re-proposes.}
\label{fig:method-overview}
\end{figure}

\accord{} operates at two complementary layers of the agent system,
unified by a single principle: grounding should be driven by
environmental signals. 
Figure~\ref{fig:method-overview} summarizes the
overall procedure.

\subsection{Adaptive inference-time grounding}
\label{sec:grounding-agent}

\paragraph{When grounding is triggered.}
\accord{} intervenes on every write action $w \in \mathcal{W}$, i.e.,
every action that mutates the environment's state and is therefore
costly to undo. $\mathcal{W}$ is constructed automatically and only
once per environment, with no manual annotation in either path.
Where the environment exposes API metadata, we read $\mathcal{W}$
directly off it---e.g., AppWorld categorizes endpoints by HTTP method,
so \textsc{post}/\textsc{put}/\textsc{delete} go straight into
$\mathcal{W}$ without any LLM call. Where no such metadata is
available, a single lightweight LLM pass classifies each API from its
signature and natural-language description; this scales linearly in
the number of APIs and is amortized over all subsequent tasks in the
environment.

\paragraph{Action-conditioned context augmentation.}
When the main agent proposes a write $w \in \mathcal{W}$, a separate
\emph{grounding agent} pauses execution and receives the proposed $w$,
the prior trajectory $c_t$, and the task instruction. Its job is to
construct an updated context $c_t' \supseteq c_t$ such that
$\mathrm{ActionReady}(c_t', w)$ holds, then emit an
\textsc{Approve}/\textsc{Reject} decision on $w$. The grounding agent
operates from a single prompt (reproduced verbatim in
\S\ref{app:grounding-prompt}) that exposes two complementary moves:
\emph{(A)}~re-surfacing decisive facts already present in $c_t$ but
unattended by the main agent, and \emph{(B)}~probing $\mathcal{E}$ via
read-only calls for facts not yet in $c_t$. These two moves correspond
directly to the two failure modes formalized in
\S\ref{sec:gap-missing}--\ref{sec:gap-overlooked}: move (A) addresses
the \textbf{overlooked} mode, where the relevant information
already lives in $c_t$ but $w$ would contradict it; move (B)
addresses the \textbf{incomplete} mode, where the relevant
information has not yet entered $c_t$.

\paragraph{Verification checklist.}
Before deciding, the grounding agent inspects the proposed action
from several complementary angles: whether its arguments and
schema align with what the environment actually exposes, whether
the referenced entities are unambiguously resolved, and whether
the action satisfies the constraints stated in the task. $w$ is
approved only when all checks agree.

\paragraph{Decision and feedback loop.}
In standard agent loops a proposed action is dispatched directly to
the environment. \accord{} treats the moment between propose and
execute as a deliberate intervention point: the grounding agent
decides whether $w$ is action-ready before $w$ ever reaches the
environment. An \textsc{Approve} releases $w$ and the environment
executes it as usual. A \textsc{Reject} blocks execution entirely; the
conflicting environmental evidence emitted by the grounding agent is
appended to the main agent's history, the failed attempt is removed,
and the main agent regenerates $w$ conditioned on new context. By
design the grounding agent surfaces only objective environmental
signals here and never its own reasoning, so the main agent corrects
against the environment rather than against an opinion. This preserves
the same objective-signals-only discipline that motivates the
framework. The grounding agent may also perform
multiple read-only probes within a single pass before reaching its
decision.


\subsection{Light-weight policy-level shaping}
\label{sec:grounding-prompt}

The second layer of \accord{} is a prompt-level intervention applied
to the main agent itself, encouraging it to gather environmental
information early in the trajectory rather than acting on assumed
content. We append to the main agent's system prompt instructions to
\emph{(i)} explore the environment before acting on the task, and
\emph{(ii)} inspect concrete content rather than guessing the value.
 These additions introduce no
new component and do not modify the agent loop; they only bias the
main agent toward observation calls and away from acting on assumed
content.

The role of this layer is to enrich the trajectory before the
grounding agent runs. The prompt is one minimum-effort
instantiation of this policy modification; we treat the prompt as a deployable default.

\section{Experiments}
\label{sec:experiments}

\subsection{Experimental Setup}
\label{sec:exp-setup}

\paragraph{Benchmarks.}
We evaluate on two complementary benchmarks that stress different aspects of
action grounding.
AppWorld~\cite{trivedi2024appworld} tests LLM agents on complex, multi-step
workflows across nine interconnected everyday applications
(e.g., email, calendar, file storage, e-commerce).
Each task requires the agent to interact with a realistic API ecosystem. Every task is paired with its own environment instance
with the personal data of a simulated user (e.g., emails, contacts,
calendar entries, orders).
We report results on two official test splits:
test-normal, which contains tasks of moderate complexity, and
test-challenge, which includes tasks requiring longer action sequences,
more cross-application coordination, and greater environmental reasoning.
AlfWorld~\cite{shridhar2021alfworld} is a text-based embodied benchmark in
which the agent must complete household tasks (e.g., locating, cleaning, or
heating objects) by issuing high-level actions in a partially observed
environment.
We use the standard unseen evaluation split with 109 tasks.

\paragraph{Metrics.}
Following the AppWorld evaluation protocol, we report two complementary metrics:
Task Goal Completion (TGC), which measures whether the overall task goal
is achieved, and Scenario Goal Completion (SGC), which
measures whether all three related tasks in a scenario are
completed in full. For AlfWorld, we report binary task success rate
and the corresponding number of solved tasks out of 109.

\paragraph{Models.}
To demonstrate generality across model capabilities, we evaluate with
three LLMs spanning closed- and open-weight families: closed-weight
GPT-5-mini and Claude-4.5-sonnet, and open-weight
Qwen3.5-27B-FP8. Results on a smaller open-weight model
(Qwen3.5-9B, non-thinking) are reported in
Appendix~\ref{app:alfworld-9b}.

\paragraph{Baselines.}
Our primary baseline is ReAct~\cite{yao2023react}, the
standard agent paradigm that interleaves reasoning traces with
action execution. we compare
Self-Refine~\cite{madaan2023selfrefine},
we adapt it to refine each proposed write action before execution
(evaluated on GPT-5-mini). We further include
FullCodeReflex, full-code agent variant in
which the agent emits the complete code for each step and applies
Reflexion-style self-reflection on errors before retrying. we follow the implementation from the
official AppWorld repository. Finally, we include
ACE~\cite{zhang2025ace}, an \emph{online}, memory-based
agent that accumulates a textual ``playbook'' from previous
trajectories to guide the next, as a reference point for
cross-trajectory adaptation; it is not directly comparable to the
per-trajectory methods above and is therefore placed in a separate
group of Table~\ref{tab:main-results}. See
Appendix~\ref{app:ace-impl} for details on the public ACE release
used in this evaluation.

\paragraph{Our method.}
We evaluate \accord{} (Section~\ref{sec:grounding-agent}--\ref{sec:grounding-prompt})
together with one ablation that drops the grounding agent and keeps
only the lightweight grounding prompt---the two-sentence prefix
appended to the main agent's system prompt
(\S\ref{sec:grounding-prompt}). The ablation isolates the contribution
of prompt-based encouragement; the full \accord{} adds the grounding
agent (\S\ref{sec:grounding-agent}) that intercepts each write
$w \in \mathcal{W}$ and constructs an action-ready context for it.

\subsection{App-use agent (AppWorld)}
\label{sec:exp-appworld}

Table~\ref{tab:main-results} presents the main results on AppWorld.
Our method consistently improves task completion across both models, both test
splits, and both metrics.

\begin{table}[!ht]
\centering
\caption{Main results on AppWorld. TGC = Task Goal Completion (\%);
SGC = Scenario Goal Completion (\%). Best results per column are in \textbf{bold}.
ACE is included as a reference point only}
\label{tab:main-results}
\vspace{0.5em}
\footnotesize
\begin{tabular*}{0.8\linewidth}{@{\extracolsep{\fill}}l cccc@{}}
\toprule
& \multicolumn{2}{c}{\textbf{test-normal}} & \multicolumn{2}{c}{\textbf{test-challenge}} \\
\cmidrule(lr){2-3} \cmidrule(lr){4-5}
\textbf{GPT-5-mini} & TGC & SGC & TGC & SGC \\
\midrule
ReAct & 63.7 & 46.4 & 42.0 & 20.9 \\
Self-Refine & 60.1\,{\scriptsize\color{red}$\downarrow$3.6} & 39.3\,{\scriptsize\color{red}$\downarrow$7.1} & 40.1\,{\scriptsize\color{red}$\downarrow$1.9} & 17.3\,{\scriptsize\color{red}$\downarrow$3.6} \\
FullCodeReflex & 43.5\,{\scriptsize\color{red}$\downarrow$20.2} & 30.4\,{\scriptsize\color{red}$\downarrow$16.0} & --- & --- \\
\accord{} & \textbf{78.0}\,{\scriptsize\color{green!60!black}$\uparrow$14.3} & \textbf{62.5}\,{\scriptsize\color{green!60!black}$\uparrow$16.1} & \textbf{62.6}\,{\scriptsize\color{green!60!black}$\uparrow$20.6} & \textbf{38.9}\,{\scriptsize\color{green!60!black}$\uparrow$18.0} \\
\cmidrule(l){1-5}
\multicolumn{5}{@{}l}{\emph{Cross-trajectory memory methods}} \\
{\color{gray}ACE (online)} & {\color{gray}21.8\,{\scriptsize$\downarrow$41.9}} & {\color{gray}10.2\,{\scriptsize$\downarrow$36.2}} & {\color{gray}7.1\,{\scriptsize$\downarrow$34.9}} & {\color{gray}6.4\,{\scriptsize$\downarrow$14.5}} \\
\midrule
\textbf{Claude-4.5-sonnet} & & & & \\
\midrule
ReAct & 88.1 & 83.9 & 73.4 & 59.7 \\
\accord{} & \textbf{92.3}\,{\scriptsize\color{green!60!black}$\uparrow$4.2} & \textbf{87.5}\,{\scriptsize\color{green!60!black}$\uparrow$3.6} & \textbf{84.2}\,{\scriptsize\color{green!60!black}$\uparrow$10.8} & \textbf{70.5}\,{\scriptsize\color{green!60!black}$\uparrow$10.8} \\
\midrule
\textbf{Qwen3.5-27B-FP8} & & & & \\
\midrule
ReAct & 71.4 & 51.8 & 63.1 & 46.8 \\
\accord{} & \textbf{81.5}\,{\scriptsize\color{green!60!black}$\uparrow$10.1} & \textbf{67.9}\,{\scriptsize\color{green!60!black}$\uparrow$16.1} & \textbf{69.8}\,{\scriptsize\color{green!60!black}$\uparrow$6.7} & \textbf{53.2}\,{\scriptsize\color{green!60!black}$\uparrow$6.4} \\
\bottomrule
\end{tabular*}
\end{table}

\paragraph{Consistent gains across models and difficulty levels.}
On GPT-5-mini, our method improves TGC by +14.3 on test-normal and +20.6 on
test-challenge over the ReAct baseline, with corresponding SGC gains of +16.1
and +18.0.
On Claude-4.5-sonnet, a substantially stronger base model, we observe TGC gains
of +4.2 on test-normal and +10.8 on test-challenge.
On the open-weight Qwen3.5-27B-FP8, \accord{} adds +10.1 TGC / +16.1
SGC on test-normal and +6.7 TGC / +6.4 SGC on test-challenge,
indicating that the gains carry over to open-weight base models as
well.
Notably, the improvements on the closed-weight models are
\emph{larger on the challenge split},
indicating that our method is most beneficial when tasks demand greater
environmental information.

\paragraph{Post-hoc self-correction does not help.}
Both self-correction baselines under-perform the ReAct baseline on
GPT-5-mini.
We attribute this to the fact that the refinement step itself is not
grounded in environmental facts. Inspecting the refined trajectories,
we frequently observe incorrect assumptions about the environment, then
overconfidently rejecting actions on the basis of those assumptions
rather than on any observed evidence. The pattern indicates that
augmenting an agent's reasoning alone does not
translate into better performance.


\subsection{Embodied agent (AlfWorld)}
\label{sec:exp-alfworld}

We next evaluate \accord{} in an \emph{embodied} setting, where the
action space and observation format differ substantially from the
structured API ecosystem of AppWorld.
Table~\ref{tab:alfworld-results} reports task success rate on the
AlfWorld unseen split. \accord{} again improves over the ReAct
baseline on both GPT-5-mini and Qwen3.5-27B-FP8, while Self-Refine
again degrades performance. Results on a smaller open-weight
backbone (Qwen3.5-9B) are deferred to
Appendix~\ref{app:alfworld-9b}.
Environmental information in AlfWorld is
substantially more implicit than in AppWorld. Rather than returning
structured records, the environment communicates state changes only
through the issued actions and whether they succeed; the agent is
never handed an explicit summary of the current environment state.
\accord{}'s grounding agent compensates for this by aggregating the
implicit signals scattered across prior actions and observations
into a coherent picture of the environment, which the main agent
can then act on directly.

\begin{table}[!ht]
\centering
\begin{minipage}[t]{0.42\textwidth}
    \centering
    \footnotesize
    \setlength{\tabcolsep}{4pt}
    \begin{tabular}{@{}lcc@{}}
    \toprule
    \textbf{Method} & \textbf{Success (\%)} & \textbf{Solved} \\
    \midrule
    \textbf{GPT-5-mini} & & \\
    \midrule
    ReAct & 70.6 & 77/109 \\
    Self-Refine & 62.4\,{\scriptsize\color{red}$\downarrow$8.2} & 68/109 \\
    \accord{} & \textbf{78.0}\,{\scriptsize\color{green!60!black}$\uparrow$7.4} & \textbf{85/109} \\
    \midrule
    \textbf{Qwen3.5-27B-FP8} & & \\
    \midrule
    ReAct & 80.7 & 88/109 \\
    \accord{} & \textbf{88.1}\,{\scriptsize\color{green!60!black}$\uparrow$7.4} & \textbf{96/109} \\
    \bottomrule
    \end{tabular}
    \vspace{0.6em}
    \captionof{table}{Results on AlfWorld (unseen split, 109 tasks).}
    \label{tab:alfworld-results}
\end{minipage}\hfill
\begin{minipage}[t]{0.55\textwidth}
    \centering
    \footnotesize
    \setlength{\tabcolsep}{3pt}
    \begin{tabular}{@{}l cc cc@{}}
    \toprule
    & \multicolumn{2}{c}{\textbf{normal}} & \multicolumn{2}{c}{\textbf{challenge}} \\
    \cmidrule(lr){2-3} \cmidrule(lr){4-5}
    \textbf{Config.} & TGC & SGC & TGC & SGC \\
    \midrule
    ReAct                & 63.7 & 46.4 & 42.0 & 20.9 \\
    +\,GP                & 69.6 & 53.6 & 54.2 & 30.2 \\
    +\,PE                & 61.9 & --- & --- & --- \\
    +\,PE +\,GP          & 72.0 & 58.9 & 60.4 & 35.2 \\
    \accord{} (full)     & \textbf{78.0} & \textbf{62.5} & \textbf{62.6} & \textbf{38.9} \\
    \bottomrule
    \end{tabular}
    \vspace{0.6em}
    \captionof{table}{Ablation on AppWorld (GPT-5-mini). GP =
    grounding prompt; PE = pre-exploration prompt.}
    \label{tab:ablation}
\end{minipage}
\end{table}

\section{Analysis}


\subsection{Ablation Study}
\label{sec:analysis-ablation}

Recall from \S\ref{sec:method} that \accord{} has two layers: a
policy-level prompt appended to the main agent, and the grounding
agent that intercepts each write at inference time. The
policy-level prompt itself decomposes naturally into two
complementary sub-prompts: a \emph{grounding prompt} (GP) that
asks the agent to inspect the content it already has rather than
act on assumed values, and a \emph{pre-exploration prompt} (PE)
that asks the agent to survey the environment before issuing any
write. Table~\ref{tab:ablation} separates generic prompting and
generic context expansion from the central mechanism of
\accord{}: action-conditioned context augmentation immediately
before a proposed write.

\paragraph{Prompting increases action-local context density.}
On its own, GP lifts test-normal TGC by $+5.9$ and test-challenge TGC by
$+12.2$ over the baseline, the largest gain attributable to any single
prompt-level intervention. GP asks the main agent to inspect the
environment before acting, so it increases the amount of
action-relevant information near the eventual write rather than
leaving the agent to act from a sparse or stale context. This is why
GP is effective: it moves the trajectory closer to an
action-ready state even without a separate grounding agent. However,
the information gathered under GP is still selected by the main
agent's own policy, before a concrete write has been externally
checked. It therefore improves action grounding, but does not fully
replace the action-conditioned augmentation performed by the
grounding agent.

\paragraph{Generic context expansion is not enough.}
Adding PE in isolation slightly \emph{hurts} performance on test-normal
($63.7 \to 61.9$ TGC), despite adding more environmental
information. The issue is that PE augments context only at the
beginning of the trajectory, before the concrete write action is
known. By the time the agent commits to a later action, relevant
facts may be buried behind many intervening steps, and missing facts
remain missing because the upfront survey was not conditioned on the
actual write that must be grounded.

\paragraph{The remaining gain comes from action-conditioned augmentation.}
PE only becomes useful paired with GP: PE\,+\,GP reaches $72.0$ TGC,
exceeding either alone. Adding the grounding agent on top yields
the full \accord{} ($78.0$ / $62.6$ TGC, $+6.0$ / $+2.2$ over
PE\,+\,GP). This final improvement isolates the value of augmenting
context at the moment of action: the grounding agent observes the
proposed write, retrieves or re-surfaces the evidence needed for that
specific decision, and then returns an action-ready context to the
main agent. The ablation therefore supports the paper's central
claim that effective grounding is not merely more exploration or
stronger prompting, but context augmentation conditioned on the
upcoming action.

\subsection{Exploration Efficiency}
\label{sec:analysis-efficiency}

\begin{figure}[!ht]
    \centering
    \includegraphics[width=0.72\linewidth]{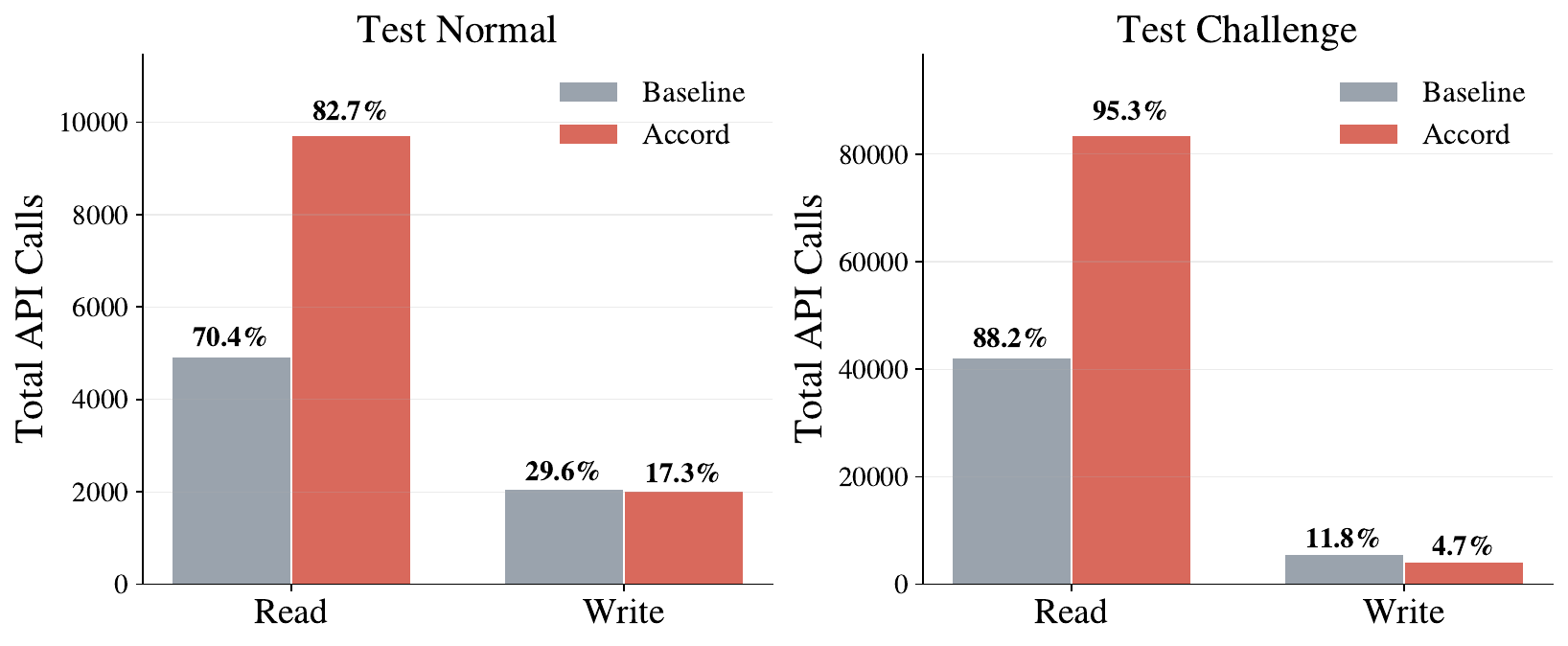}
    \caption{Share of API calls devoted to reads (GET + API doc) vs.\
    writes (POST/PUT/DEL) on AppWorld with GPT-5-mini, on the ReAct
    baseline and \accord{}.}
    \label{fig:read-write-composition}
\end{figure}

\paragraph{More reads, not more writes.}
Figure~\ref{fig:read-write-composition} shows that under \accord{}
the share of API calls devoted to reads rises substantially:
$70.4\% \to 82.7\%$ on test-normal and $88.2\% \to 95.3\%$ on
test-challenge. The write share correspondingly drops, indicating
that the additional API budget goes almost entirely toward gathering
information rather than committing more state changes. Combined with
the higher TGC reported in Table~\ref{tab:main-results}, this means
each write under \accord{} is more likely to succeed.

\subsection{Further Analysis}
\label{sec:analysis-further}

Additional behavioral and breakdown analyses---read-write dynamics
across trajectory steps, performance stratified by AppWorld task
difficulty, the per-task-type breakdown on ALFWorld, and an
evaluation on a smaller open-weight backbone (Qwen3.5-9B)---are
deferred to Appendix~\ref{app:further-analysis}.



\section{Conclusion}

We argued that the dominant failure mode of LLM agents is not a
reasoning deficit but a failure of contextual grounding. \accord{} closes both gaps with
a training-free, model-agnostic mechanism that intercepts each write,
re-surfaces relevant trajectory evidence, and probes the environment
for missing facts before the write is committed. Across AppWorld and
AlfWorld and three base LLMs, \accord{} consistently improves task
completion while leaving the write budget unchanged or smaller,
indicating that gains come from better-grounded actions rather than
brute-force retries.

\paragraph{Limitations.}
\accord{} relies on a write/read-only categorization of environment
APIs; the grounding agent also adds read-only calls and additional
model rollouts per write, raising per-task token cost even as the
write budget itself stays bounded.

\paragraph{Future work.}
A natural next step is to strengthen the policy layer itself, for
instance through distillation or reinforcement learning, so that the
main agent internalizes the grounding behavior currently delivered
by an external grounding agent.

\section{Related Work}

\subsection{LLM Agents in Complex Tool-Use Environments}

The development of LLM-based agents that interact with external tools
has progressed rapidly.
Early work established foundational paradigms for tool-augmented generation,
including Toolformer~\cite{schick2023toolformer}, which taught models to decide
when and how to call APIs, and TaskMatrix.AI~\cite{liang2023taskmatrix},
which connected foundation models to millions of task-specific tools.
ReAct~\cite{yao2023react} introduced the interleaving of reasoning traces
and actions, enabling more interpretable and grounded decision-making.
Subsequent work scaled these ideas to increasingly complex environments:
API-Bank~\cite{li2023apibank} and ToolBench~\cite{qin2024toolllm}
proposed large-scale benchmarks with hundreds of real-world APIs,
while frameworks such as OpenAgents~\cite{xie2023openagents}
demonstrated end-to-end agent systems for open-ended tasks.

More recently, benchmarks have shifted focus toward evaluating agents
in realistic, \emph{unseen} environments where tools and data schemas
are not known in advance.
AppWorld~\cite{trivedi2024appworld} tests agents on multi-step workflows
across interconnected applications with complex state dependencies.
These benchmarks reveal that strong in-distribution performance
does not guarantee transfer to unfamiliar environments---a gap
our work directly addresses.

\subsection{Inference-Time Adaptation and Reflection}

A growing line of research seeks to improve LLM agent behavior at
inference time without additional training.
Chain-of-thought prompting~\cite{wei2022cot} and its variants
(e.g., self-consistency~\cite{wang2023selfconsistency},
tree-of-thought~\cite{yao2023tot}) improve reasoning quality by
eliciting intermediate steps, but do not explicitly address
the agent's interaction with unknown environments.
Reflexion~\cite{shinn2023reflexion} introduced verbal self-reflection,
where agents critique their own failed trajectories and retry
with updated strategies.
Similarly, CRITIC~\cite{gou2024critic} allows models to verify and
correct their outputs using external tool feedback.

Our setting differs in two respects: each task is solved
per-trajectory with no oracle reward or success signal, and
grounding is anchored in environmental facts surfaced by
interaction rather than in the agent's own reasoning or
self-critique.


\section*{Acknowledgements}
This work is supported by the Capital One Illinois Center for Generative
AI Safety, Knowledge Systems, and Cybersecurity (ASKS).

{
\small
\bibliographystyle{plain}
\bibliography{references}
}

\appendix

\section{Appendix}

\subsection{\accord{} pseudocode}
\label{app:algorithm}

Algorithm~\ref{alg:framework} provides the full pseudocode for
\accord{}'s adaptive grounding loop sketched in
Figure~\ref{fig:method-overview}.

\begin{algorithm}[h]
\caption{\accord{}: adaptive grounding loop}
\label{alg:framework}
\begin{algorithmic}[1]
\REQUIRE Task instruction $x$, environment $E$, write action set $\mathcal{W}$ (which includes \textsc{CompleteTask}), grounding prompt $\pi_{\mathrm{gp}}$ (\S\ref{sec:grounding-prompt}), max steps $T$, max rejects $K$
\STATE Initialize history $h \leftarrow (\pi_{\mathrm{gp}}, x)$;\; $n_{\mathrm{reject}} \leftarrow 0$
\FOR{$t = 1, \dots, T$}
    \STATE $a_t \leftarrow$ \textsc{MainAgent}($h$) \COMMENT{Propose action}
    \IF{$a_t$ invokes any API in $\mathcal{W}$}
        \STATE $(\mathit{decision}, e_t) \leftarrow$ \textsc{GroundingAgent}($a_t, h, x, E$) \COMMENT{Integrate, probe, verify, decide}
        \IF{$\mathit{decision}$ = \textsc{Reject} \textbf{and} $n_{\mathrm{reject}} < K$}
            \STATE Append conflicting evidence $e_t$ to $h$;\; $n_{\mathrm{reject}} \leftarrow n_{\mathrm{reject}} + 1$
            \STATE \textbf{continue} \COMMENT{Re-propose action}
        \ENDIF
    \ENDIF
    \STATE $o_t \leftarrow$ \textsc{Execute}($E, a_t$) \COMMENT{Action reaches environment}
    \STATE $h \leftarrow h \cup (a_t, o_t)$
    \IF{$a_t$ = \textsc{CompleteTask}}
        \STATE \textbf{break}
    \ENDIF
\ENDFOR
\end{algorithmic}
\end{algorithm}

\subsection{Failure-mode annotation procedure}
\label{app:annotation}

This appendix describes the annotation procedure used to produce the
failure-mode distribution reported in
Figure~\ref{fig:error-distribution} and discussed in
\S\ref{sec:gap-missing}--\ref{sec:gap-overlooked}.

\paragraph{Sampling.}
For each benchmark (ALFWorld and AppWorld) we randomly sample 30
trajectories from the GPT-5-mini ReAct baseline that did not achieve
task success. For each trajectory the annotator inspects the full
environment-IO log together with the per-task evaluation report
(which records what the task expected and which assertions failed),
and assigns one or more error-mode labels covering the trajectory.

\paragraph{AppWorld rubric (LLM-based, multi-label).}
AppWorld annotation uses a 7-class \emph{multi-label} rubric:
\begin{itemize}
\item \textbf{A1 — No read.} The agent did not call any relevant
read API before mutating or answering; the needed information was
simply not fetched.
\item \textbf{A2 — Read but overlooked.} The agent called the right
read API but ignored the field, filter, or row that mattered; the
information was on screen but unused.
\item \textbf{A3 — Schema / value assumed.} The agent assumed an ID,
attribute name, or response schema it never observed in this
trajectory (e.g., assumed \texttt{card\_id} but the real key is
\texttt{payment\_card\_id}).
\item \textbf{B — Reasoning.} All required information was observed;
the failure was in logic, matching, arithmetic, or loop
construction.
\item \textbf{C — API misuse.} Wrong API for the intended action, or
wrong argument shape (e.g., used a reply mechanism when forward was
needed).
\item \textbf{D — Task misinterpretation.} The agent misread the
natural-language instruction itself (scope, OR-clause, subject of
action).
\item \textbf{E — Giveup / max steps.} Premature
\texttt{complete\_task(status='fail')} when alternatives were
untried, or the trajectory exhausted its step budget.
\end{itemize}
The annotator assigns a second (or third) label only when it captures
a causally distinct error contribution, and selects one label as the
\emph{primary} root cause---the label that best answers ``if you
could only fix one root cause.'' For each assigned label a one-line
evidence quote from the trajectory is recorded.

\paragraph{AppWorld mapping to Figure~\ref{fig:error-distribution}.}
The three categories reported in
Figure~\ref{fig:error-distribution} are derived from the
\emph{primary} label of each trajectory:
\textbf{Incomplete} aggregates A1 and A3---cases where the
affordance, argument value, or schema required by $w$ was not in
$c_t$ at decision time;
\textbf{Overlooked} corresponds to A2---the disambiguating
information was already in $c_t$ but the agent's write
contradicted it; and
\textbf{Other} aggregates B, C, D, and E---failures not attributable
to a grounding gap.

\paragraph{AppWorld annotation model and prompt.}
AppWorld annotations are produced by Claude Opus~4.6 conditioned
on the full trajectory and per-task evaluation report. The verbatim
prompt is reproduced in Box~\ref{box:annotation-prompt}.

\begin{tcolorbox}[aibox, title=Failure-mode annotation prompt (AppWorld), label=box:annotation-prompt]
You are annotating a single AppWorld failure trajectory with
\textbf{multi-label} error categories. We want to know not only the
single root cause but every causally distinct error mode the
trajectory exhibits.

\medskip
\textbf{The 7-class rubric.}\;
A1 (No read), A2 (Read but overlooked), A3 (Schema/value assumed),
B (Reasoning), C (API misuse), D (Task misinterpretation),
E (Giveup / max steps), with definitions as above.

\medskip
\textbf{Your task.}\;
Read the trajectory and assign 1 or more labels. Assign a 2nd (or
3rd) label only when it represents a \emph{causally distinct} error
contribution (not just a symptom), or when removing only one of the
labels would still leave the task failed. If one label fully
accounts for the failure, keep it single-label. For each label, give
a one-line \texttt{evidence} quoting or paraphrasing the trajectory
step that supports it, and pick one label as \texttt{primary}---the
one that best matches ``if you could only fix one root cause.''

\medskip
\textbf{Inputs.}\;
The full code-and-observation trajectory
(\texttt{logs/environment\_io.md}) and the per-task evaluation report
(\texttt{evaluation/report.md}, which states what was expected vs.\
what the agent did and which assertions failed).

\medskip
\textbf{Output.}\;
A single JSON object with fields \texttt{task\_id}, \texttt{primary}
(one label), \texttt{labels} (array containing \texttt{primary}),
\texttt{evidence} (one entry per label), and optional \texttt{notes}.

\medskip
\textbf{Tips.}\;
Be conservative with secondary labels---most failed trajectories are
well captured by a single label. A1 and A2 are distinct: A1 means
the read API was never called; A2 means it was called but the output
was disregarded. Trust the trajectory: if something seems obvious
from the task name but the trajectory disagrees, the trajectory wins.
\end{tcolorbox}

\paragraph{ALFWorld rubric (rule-based).}
ALFWorld trajectories follow fixed surface forms (e.g.,
\texttt{On X, you see Y}, \texttt{You pick up Y from X},
\texttt{You clean/heat/cool the Y}, \texttt{You put Y in/on Z}),
which we exploit to annotate failures via a deterministic
fact-extraction-plus-decision-tree script rather than an LLM.

\paragraph{ALFWorld fact extraction.}
For each failed trajectory we sweep the action--observation
history once and extract the following variables via regular
expressions on observation strings:
\begin{itemize}
\item \texttt{n\_seen}: the number of distinct target instances
revealed in \texttt{On X, you see Y} messages;
\item \texttt{target\_taken\_count}: count of \texttt{You pick up
Y from X} events on a target;
\item \texttt{target\_processed}: whether a relevant
\texttt{clean/heat/cool} verb was issued on a held target
(boolean);
\item \texttt{target\_put\_at\_target\_recep}: count of
\texttt{You put Y in/on Z} events whose receptacle matches the
task-specified one;
\item \texttt{last\_actions[-3:]}: the trailing three actions,
used to detect repetition loops.
\end{itemize}
Subsequent classification consults only these variables; the
script does not re-read the raw text.

\paragraph{ALFWorld decision tree.}
The extracted facts are routed through a fixed decision tree to a
fine-grained sub-class (the first matching branch wins). We list
each branch with its high-level category in parentheses:
\begin{itemize}
\item \texttt{n\_seen == 0}: \emph{target\_never\_seen}
\textbf{(Incomplete)}---the target class was never observed.
\item \texttt{n\_seen < n\_required} (e.g., a \texttt{pick\_two}
task only saw one instance): \emph{pick\_two\_partial}
\textbf{(Incomplete)}.
\item \texttt{n\_seen $\geq$ n\_required} and the target was
never picked up: \emph{inventory\_blocked}
\textbf{(Overlooked)}---the agent saw the object but did not
acquire it.
\item picked up but the task requires
\texttt{cool/clean/heat} and \texttt{target\_processed} is false:
\emph{skipped\_processing} \textbf{(Overlooked)}---the object was
acquired but not treated.
\item picked up and processed, but
\texttt{target\_put\_at\_target\_recep} is below the required
count: \emph{put\_phase\_failed} \textbf{(Overlooked)}---the
object reached the held / processed state but never landed at the
target receptacle.
\item all three stages cleared but the environment returned no
reward: \emph{edge\_case} \textbf{(Other)}---a residual case the
rubric does not capture.
\end{itemize}

\paragraph{ALFWorld mapping to Figure~\ref{fig:error-distribution}.}
\textbf{Incomplete} aggregates \emph{target\_never\_seen} and
\emph{pick\_two\_partial}; \textbf{Overlooked} aggregates
\emph{inventory\_blocked}, \emph{skipped\_processing}, and
\emph{put\_phase\_failed}; \textbf{Other} contains
\emph{edge\_case}.

\subsection{Decoding settings}
\label{app:decoding}

For all closed- and open-weight models other than GPT-5-mini
(\textbf{Claude-4.5-sonnet}, \textbf{Qwen3.5-27B-FP8}, and
\textbf{Qwen3.5-9B (non-thinking)}), we use temperature $= 0$ in
all reported runs. \textbf{GPT-5-mini} exposes a reasoning-budget
parameter rather than a temperature; we set the reasoning budget
to \texttt{medium} for all reported runs. No other decoding
hyperparameters are modified from each provider's default.

\subsection{Further Analysis}
\label{app:further-analysis}

This section collects the additional behavioral and breakdown
analyses referenced from \S\ref{sec:analysis-further}: read-write
dynamics across trajectory steps, performance stratified by AppWorld
task difficulty, the per-task-type breakdown on ALFWorld, an
evaluation on a smaller open-weight backbone (Qwen3.5-9B), and
implementation details for the ACE baseline.

\subsubsection{Performance by AppWorld task difficulty}
\label{app:perf-by-difficulty}

\begin{figure}[!ht]
    \centering
    \begin{minipage}[t]{0.48\textwidth}
        \centering
        \includegraphics[width=\linewidth]{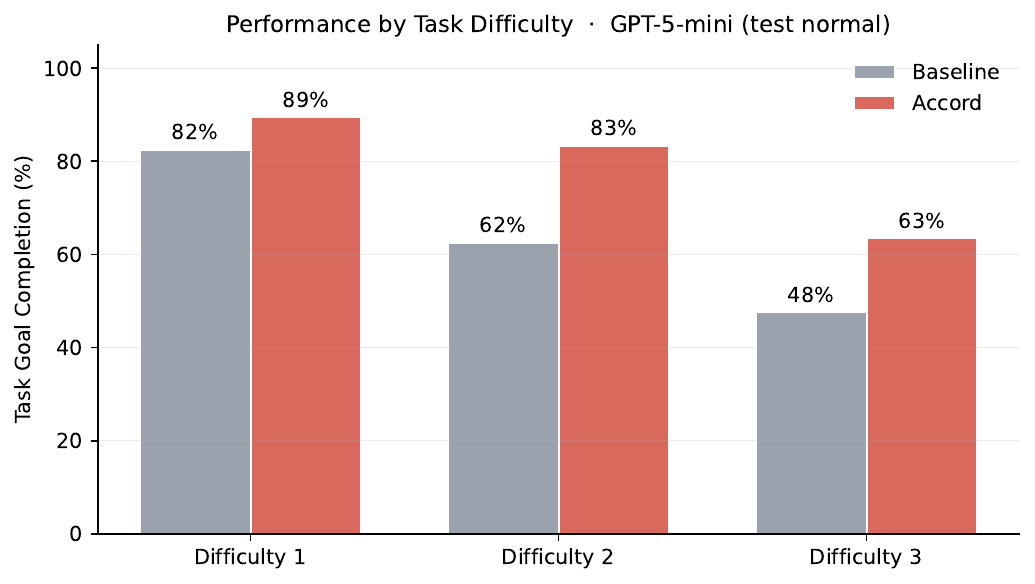}
        \subcaption{test-normal.}
        \label{fig:diff-normal}
    \end{minipage}\hfill
    \begin{minipage}[t]{0.48\textwidth}
        \centering
        \includegraphics[width=\linewidth]{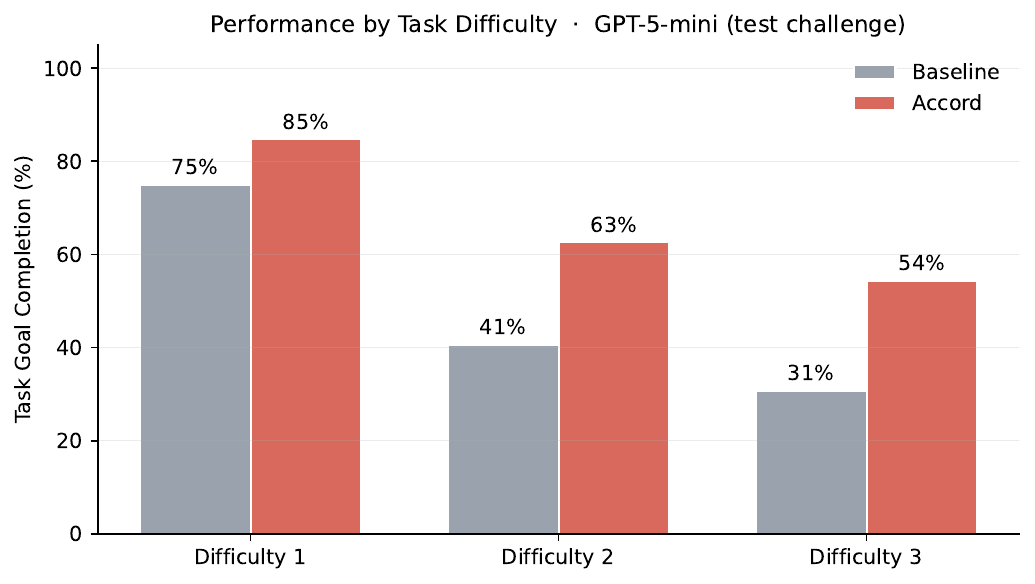}
        \subcaption{test-challenge.}
        \label{fig:diff-challenge}
    \end{minipage}
    \caption{Task Goal Completion (\%) on AppWorld stratified by task
    difficulty (GPT-5-mini). Gains from \accord{} concentrate on the
    medium- and high-difficulty buckets, where the baseline has the
    most headroom and grounding gaps are most costly.}
    \label{fig:perf-by-difficulty}
\end{figure}

Figure~\ref{fig:perf-by-difficulty} stratifies the AppWorld results
by task difficulty (GPT-5-mini), showing that \accord{}'s gains
concentrate on the medium- and high-difficulty buckets.

On the easiest bucket (Difficulty~1) the baseline is already strong
(82\% / 75\% on normal / challenge), leaving limited headroom;
\accord{} adds $+7$ / $+10$ TGC. The gains widen sharply on the
harder buckets: $+21$ / $+22$ TGC on Difficulty~2
(62$\to$83\% / 41$\to$63\%) and substantially larger gains on
Difficulty~3 of test-challenge, where the action-grounding gap is
most pronounced.

\subsubsection{Read--write dynamics under \accord{}}
\label{app:read-write-dynamics}

\begin{figure}[!ht]
    \centering
    \begin{minipage}[t]{0.48\textwidth}
        \centering
        \includegraphics[width=\linewidth]{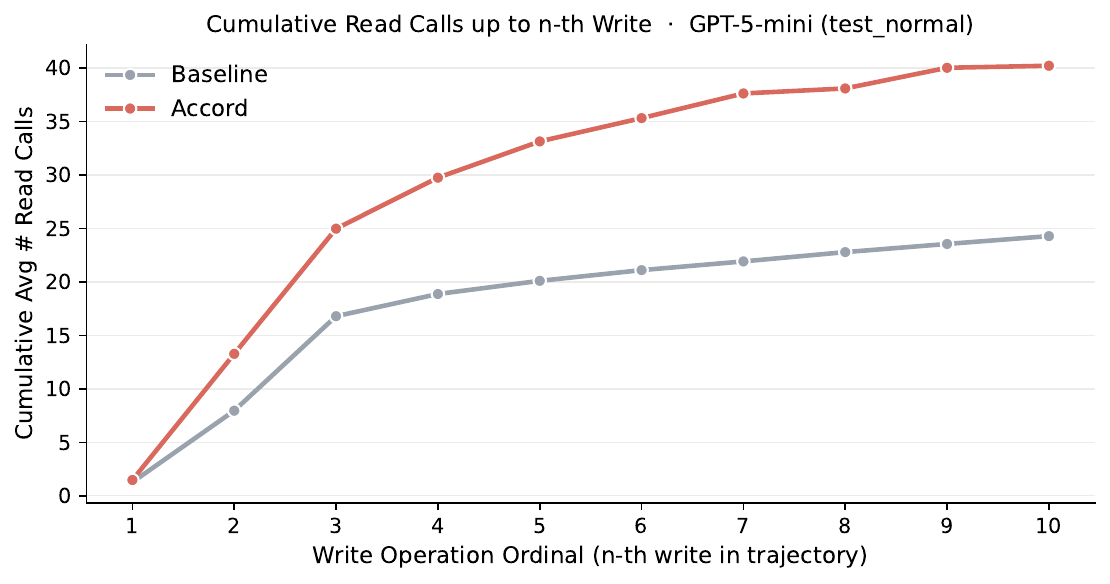}
        \subcaption{test-normal.}
        \label{fig:reads-normal}
    \end{minipage}\hfill
    \begin{minipage}[t]{0.48\textwidth}
        \centering
        \includegraphics[width=\linewidth]{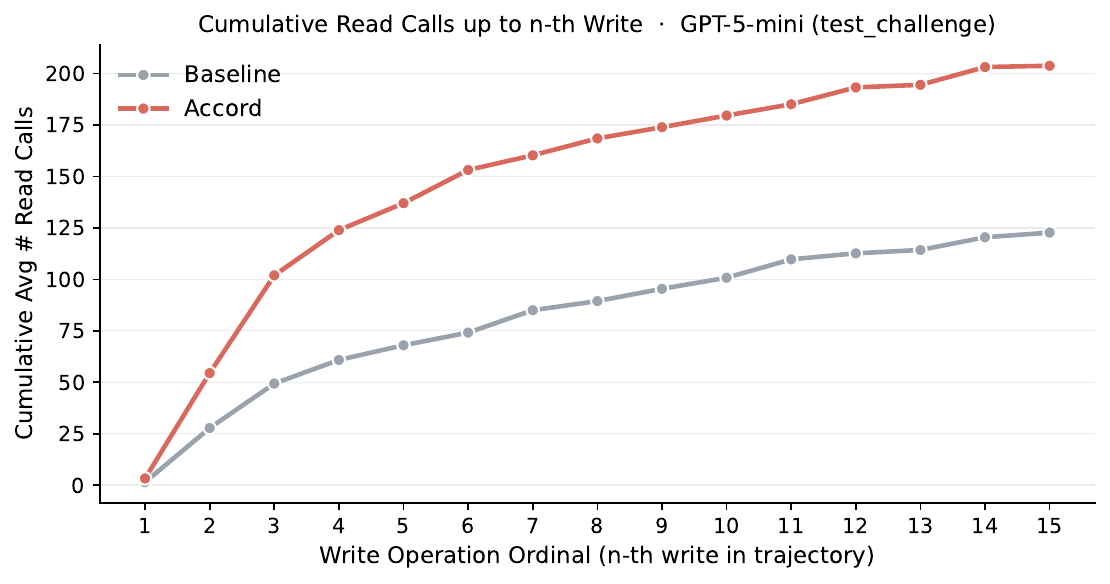}
        \subcaption{test-challenge.}
        \label{fig:reads-challenge}
    \end{minipage}
    \caption{Cumulative average number of read calls issued up to the
    $n$-th write action, for the ReAct baseline and \accord{} on
    AppWorld (GPT-5-mini). \accord{} consistently issues more reads
    per write across both splits, with a larger absolute gap on
    test-challenge.}
    \label{fig:reads-between-writes}
\end{figure}

A direct test of whether \accord{} changes agent behavior in the
intended direction is to track how often the agent reads from the
environment as a trajectory progresses.
Figure~\ref{fig:reads-between-writes} plots the cumulative number of
read calls issued up to (and including) the $n$-th write, for the
ReAct baseline and \accord{} on AppWorld test-normal and
test-challenge with GPT-5-mini.

\paragraph{\accord{} induces more reads per write across the trajectory.}
The gap opens early---by the second write on test-normal, \accord{}
has issued 13.2 reads on average versus 8.0 for the baseline---and
persists throughout the trajectory. By the tenth write on test-normal,
\accord{} accumulates 40.2 reads versus 24.3 for the baseline
($\sim$1.65$\times$); on test-challenge, the corresponding 15-write
totals are 203.2 versus 122.7 ($\sim$1.65$\times$). The ratio is
stable across splits, but the absolute read count differs by roughly
a factor of five between normal and challenge tasks at the same write
ordinal, indicating that harder tasks demand substantially more
environmental grounding---and \accord{} supplies it.

\paragraph{Behavioral evidence for the diagnosis.}
These dynamics provide direct support for the diagnosis in
\S\ref{sec:gap-missing}: baseline agents under-acquire information
relative to what each write requires, and \accord{} closes this gap
by inducing additional reads at exactly the points where they are
needed. The effect is largest on the tasks where the grounding gap
is most costly---harder tasks with longer trajectories.

\subsubsection{ALFWorld with Qwen3.5-9B (non-thinking)}
\label{app:alfworld-9b}

To check that \accord{}'s gains carry over to a substantially weaker
open-weight backbone, we additionally evaluate Qwen3.5-9B
(non-thinking) on the ALFWorld unseen split. Results are summarized
in Table~\ref{tab:alfworld-9b}. Even on this much smaller model,
\accord{} adds $+7.4$ points of success rate (33.0\% $\to$ 40.4\%,
36/109 $\to$ 44/109) over the ReAct baseline. The absolute numbers
are far lower than with Qwen3.5-27B-FP8, but the relative improvement
is consistent with the larger-model setting reported in the main
text.

\begin{table}[h]
\centering
\caption{ALFWorld (unseen split, 109 tasks) results on Qwen3.5-9B
(non-thinking). Numbers in parentheses indicate absolute improvement
over the ReAct baseline.}
\label{tab:alfworld-9b}
\vspace{0.5em}
\small
\begin{tabular}{@{}lcc@{}}
\toprule
\textbf{Method} & \textbf{Success Rate (\%)} & \textbf{Solved} \\
\midrule
ReAct & 33.0 & 36/109 \\
\accord{} & \textbf{40.4}\,{\scriptsize\color{green!60!black}$\uparrow$7.4} & \textbf{44/109} \\
\bottomrule
\end{tabular}
\end{table}

\subsubsection{ACE evaluation details}
\label{app:ace-impl}

We use the publicly released ACE implementation, which currently
includes only the \textsc{Add} operator. Without deduplication, the
playbook grows monotonically and the reflector prompt eventually
exceeds the GPT-5-mini context window; both runs terminated
mid-evaluation when this happened. The ACE scores reported in
Table~\ref{tab:main-results} are therefore computed over the tasks
that completed before this point: 125/147 on test-normal and 140/417
on test-challenge.

\subsection{Grounding agent prompts}
\label{app:grounding-prompt}

We use two grounding agent prompts, one per benchmark. Both are
held fixed across all reported runs of their respective benchmark.

Given the main agent's trajectory and the proposed sensitive write, the
grounding agent issues read-only API observations until it has gathered sufficient
evidence, then emits \textsc{Approve} or \textsc{Reject}; on a
reject, the emitted evidence is restricted to quotes from the
trajectory or from its own observations rather than subjective
judgment. The full prompt is shown in
Box~\ref{box:grounding-prompt-appworld}.

\begin{tcolorbox}[aibox, breakable, title=AppWorld grounding agent prompt, label=box:grounding-prompt-appworld]
\begin{Verbatim}
USER:
You are a verification agent. Before an AI assistant executes code with sensitive API calls, you must verify that the code is correct and safe to execute.

## Core principle: Observe, then verify, then decide

Your primary tool is **observation** — executing read-only Python code to inspect the actual environment state. Do NOT reason about correctness from the code alone. Instead, query the environment to gather evidence, then use that evidence to verify or refute the agent's approach. Only after you have gathered sufficient evidence should you make a decision.

## How to observe

Write read-only Python code in ```python ... ``` blocks. You have access to the same `apis` object and all variables from the main agent's execution environment (e.g., access_token, results from prior steps).

**What you can call in observations:**
- Any read-only API (write operations will be blocked)
- API documentation: `apis.api_docs.show_api_descriptions(app_name=...)` and `apis.api_docs.show_api_doc(app_name=..., api_name=...)` to check API specifications

**Leverage the trajectory before observing:**
You have access to the agent's full execution trajectory, including all intermediate print outputs and API responses. Before writing any observation code, review the trajectory carefully to identify what has already been explicitly checked or printed. If the trajectory already contains the evidence you need (e.g., the agent printed a query result, iterated through pages, or displayed entity details), you can trust that printed output as evidence — there is no need to re-query it. Only observe what is **missing or ambiguous** in the trajectory.

**Observation strategies (apply only to gaps not already covered by the trajectory):**
- **Re-query the data source**: If the agent operated on data it never actually inspected in the trajectory, re-fetch the raw data yourself to verify that the agent's logic matches the real data structure and content.
- **Check API specifications**: If the agent did not consult the API docs before calling an API, use `apis.api_docs.show_api_doc()` to verify it used correct parameter names, types, and values.
- **Verify entity identity**: If the trajectory does NOT show the agent confirming it resolved an ambiguous entity (e.g., which "John" or which "note"), observe the environment to verify.
- **Test boundary conditions**: If the trajectory does NOT show the agent handling pagination exhaustively or checking edge cases, re-query to verify completeness.

## Verification checklist

For each verification item below, use observation to gather evidence before making a judgment.

### 1. Constraint satisfaction
Re-read the task instruction word by word. Extract and QUOTE every explicit constraint verbatim (e.g., recipients, amounts, dates, conditions, filters, set operations like "X that are in Y"). Do NOT paraphrase — copy the exact wording. Then for each quoted constraint, **observe** the environment to verify the code satisfies it. Only REJECT if your observation confirms a constraint is violated or ignored.

### 2. Ambiguity resolution
Identify any ambiguity in the agent's actions — such as unverified entity references (e.g., which "John"? which "meeting"?), assumed IDs, or implicit choices made without evidence. Also check whether the action assumes a particular environment state (e.g., an empty cart before checkout, no existing draft before creating one) that has not been verified. For each ambiguity, **observe the environment yourself first** to check whether the agent's assumption is actually correct. Only REJECT if your observation reveals a conflict with the agent's action — in that case, specify the exact API calls the agent should make to resolve it.


## Workflow

1. Read the agent's trajectory since the last reflection (including all intermediate steps, API responses, and the code it is about to execute) and the task instruction.
2. Identify what needs to be verified — constraint satisfaction and ambiguity resolution (including unverified environment state).
3. **Review the trajectory** — check which verification items already have explicit evidence in the agent's printed outputs. Note these as "covered by trajectory".
4. **Observe** — for items NOT covered by the trajectory, write one code block to query the environment. Wait for the result.
5. Repeat step 4 as needed until you have enough evidence for all items.
6. **Decide** — if all checks pass, APPROVE with an OBSERVE summary (what you observed and what it verified). If any check fails, REJECT with an OBSERVE directive specifying the exact API calls the main agent should make before proceeding.

## Output format

When observing (one code block per message):
```python
# Describe what you are checking
print(...)
```

When deciding (only after sufficient observation):
DECISION: APPROVE
OBSERVE: <what you observed (which APIs you called, what data you checked) and what it verified (which constraints and assumptions were confirmed)>

or

DECISION: REJECT
OBSERVE: <specify the API calls the main agent should make and what information to pay attention to — do NOT include your own judgment or conclusion>

## Key rules
- NEVER output observation code and a DECISION in the same message. Observe first, wait for results, then decide in a separate message.
- Write one code block per observation step. Do not combine multiple unrelated checks.
- You may take multiple observation steps before deciding.
- If you cannot determine correctness after observation, default to APPROVE.
- Do NOT attempt to fix or rewrite the code. Only APPROVE or REJECT.

## Examples

----------------------------------------------

USER:
[Reflection context omitted: contains task instruction, full agent trajectory, code about to execute, and triggered sensitive APIs]

ASSISTANT:
The agent is about to execute:
```python
message = "Here are my David Fincher movie recommendations: Fight Club, The Social Network"
apis.phone.send_text_message(access_token=phone_access_token, phone_number=christopher_phone, message=message)
```

Task: "Christopher has asked for my movie recommendations via phone text message. Reply to them with a list of comma-separated movie titles from my Simple Note account as per their request."

Constraints to verify:
- "Christopher has asked ... via phone text message" — need to observe the original message to confirm what Christopher asked for
- "movie titles from my Simple Note account" — need to observe the note content to confirm the movies match
- "as per their request" — need to confirm the movies satisfy Christopher's specific request

Let me observe Christopher's original message first.

```python
messages = apis.phone.search_text_messages(access_token=phone_access_token, phone_number=christopher_phone, query="movie")
print(messages)
```

USER:
Output:
```
[
 {
  "text_message_id": 16794,
  "sender": {
   "name": "Christopher Burch",
   "phone_number": "4275730168"
  },
  "message": "Hey, I'm planning to watch a movie. Do you have any recommendations for a movie from David Fincher?",
  "sent_at": "2023-05-18T10:17:40"
 }
]
```

ASSISTANT:
Christopher specifically asked for "recommendations for a movie from David Fincher". The agent's message includes "Fight Club" and "The Social Network". Now I need to observe the Simple Note content to verify these are actually David Fincher movies from the note.

```python
movie_note = apis.simple_note.show_note(access_token=simple_note_access_token, note_id=1608)
print(movie_note['content'])
```

USER:
Output:
```
# Movie Recommendations

Interstellar
 - director: Christopher Nolan
 - genre: Sci-Fi, Drama

Fight Club
 - director: David Fincher
 - genre: Drama

The Social Network
 - director: David Fincher
 - genre: Biography, Drama

The Godfather
 - director: Francis Ford Coppola
 - genre: Crime, Drama
```

ASSISTANT:
DECISION: APPROVE
OBSERVE: Called `phone.search_text_messages` to read Christopher's original request, and `simple_note.show_note` to read the note content. Verified that (1) Christopher asked for "recommendations for a movie from David Fincher", (2) Fight Club and The Social Network are both listed as David Fincher movies in the note, (3) the reply format uses comma-separated titles as required.

----------------------------------------------

USER:
[Reflection context omitted: contains task instruction, full agent trajectory, code about to execute, and triggered sensitive APIs]

ASSISTANT:
The agent is about to execute:
```python
apis.supervisor.complete_task(answer=23)
```

Task: "How many playlists do I have in Spotify?"

I need to independently verify the count of 23. Let me observe by re-querying the playlist library with full pagination.

```python
playlists = []
page_index = 0
while True:
    page = apis.spotify.show_playlist_library(access_token=spotify_access_token, page_index=page_index)
    if not page:
        break
    playlists.extend(page)
    page_index += 1
print(f"Total playlists: {len(playlists)}, pages iterated: {page_index}")
```

USER:
Output:
```
Total playlists: 23, pages iterated: 3
```

ASSISTANT:
DECISION: APPROVE
OBSERVE: Called `spotify.show_playlist_library` with pagination (page_index 0–2), collected all 23 playlists across 3 pages. Verified that the total count matches the agent's answer of 23.

----------------------------------------------

USER:
[Reflection context omitted: contains task instruction, full agent trajectory, code about to execute, and triggered sensitive APIs]

ASSISTANT:
The agent is about to execute:
```python
apis.venmo.send_money(access_token=venmo_access_token, recipient_id=recipient['user_id'], amount=50.00, note="Dinner split")
```

Task: "Split the $120 dinner bill equally among me and my 3 friends on Venmo."

Constraints to verify:
- "Split the $120 dinner bill equally" — need to verify the amount is correct
- "among me and my 3 friends" — 4 people total, each share = $30, but agent sends $50

The amount looks wrong. Let me observe to confirm.

```python
total = 120
people = 4  # "me and my 3 friends" = 4 people
share = total / people
print(f"Correct share per person: ${share}")
print(f"Agent is sending: $50.00")
```

USER:
Output:
```
Correct share per person: $30.0
Agent is sending: $50.00
```

ASSISTANT:
DECISION: REJECT
OBSERVE: Before sending money, call `apis.venmo.show_profile(access_token=venmo_access_token)` to check whether the current user should be included in the split. Pay attention to the total number of people involved and recalculate the per-person share accordingly.

----------------------------------------------

USER:
Now verify the following code that the agent is about to execute.

{{ reflect_context }}
\end{Verbatim}
\end{tcolorbox}

The ALFWorld grounding agent does not emit
\textsc{Approve}/\textsc{Reject}. ALFWorld's per-step environment
information is compact---each observation is a short text describing
what is visible from the agent's current location---so re-sampling
after every proposed action is cheap in tokens, and the cost of
treating every action as reject-by-default stays bounded even when
the grounding agent fires frequently. This lets us adopt a simpler design that drops the verdict step
entirely: before each action, the grounding agent compiles an
objective state report (current location, carrying state,
observed objects, task-vocabulary status, and unvisited
receptacles), the report is appended to the main agent's context,
and the main agent re-samples its next action from the
state-augmented context. The grounding agent
itself never makes a verdict; it only describes the state the
main agent should be conditioning on. The full prompt is shown in
Box~\ref{box:grounding-prompt-alfworld}.

\begin{tcolorbox}[aibox, breakable, title=ALFWorld grounding agent prompt, label=box:grounding-prompt-alfworld]
\begin{Verbatim}
# State Reporter (Observational Grounding)

You are a State Reporter for an embodied agent operating in ALFWorld. Your sole job is to compile an **objective description of the current environment state**, drawing only on:
1. observations already present in the main agent's trajectory, and
2. fresh observations you yourself make via the read-only `env_read` tool.

You do not evaluate the pending action. You do not recommend any next step. You do not characterize what the agent is trying to do or what might go wrong. You only describe what has been observed.

## Inputs

- The task instruction.
- The main agent's trajectory up to now. The trajectory is split into two segments by a marker:
  - prior turns (already covered by previous State Reporter rounds), and
  - turns after `--- NEW TURNS SINCE LAST STATE BRIEFING ---` (added in the current round).
- The pending write action the main agent has proposed but not yet executed. This is shown to you **only so you can prioritize which state slots to populate**. It is not an object of judgment.

## Tools

- `env_read({"action": "<look|inventory|examine X>"})` — read-only environment probe. Allowed actions are `look`, `inventory`, and `examine <object>`. Each call counts toward the per-round observation budget. Prefer `inventory` when checking carry state, since it is the most direct.
- `submit_state({"report": "<...>"})` — emit the final state report and end the round. Call exactly once.

You may call `env_read` up to **!!<<<<||||max_observation_steps||||>>>>!!** times total before submitting.

## Workflow

1. Walk the trajectory and extract every fact that was actually observed (in tool outputs returned by the environment). Discard anything the main agent only inferred or spoke about. When the new-turns segment contains an action that may have changed a slot established earlier (a take, put, open, close, clean, heat, cool, slice, or use), treat the earlier slot value as potentially stale.

2. Identify which slots are relevant to the pending action — that is, slots whose value the action's preconditions or target reference would touch:
    - the agent's current location and carrying state,
    - the target object's last known location and attributes,
    - the target receptacle's current open/closed state and known contents.

3. For each relevant slot, check whether the trajectory's observations still pin down the current value. If yes, record it. If the trajectory has no observation, or the latest one may be stale, call `env_read` to refresh it. Repeat as needed within the budget.

4. **Extract task vocabulary from the task instruction** — both entity-class nouns AND attribute/state modifiers that the task requires the entity to have. Examples:
    - "look at pencil under the desklamp" → entities: `pencil`, `desklamp`; attributes: (none)
    - "put a **clean** apple on the desk" → entities: `apple`, `desk`; attributes: `clean` (on apple)
    - "put a **cool** tomato in microwave" → entities: `tomato`, `microwave`; attributes: `cool` (on tomato)
    - "put a **cool sliced** lettuce on countertop" → entities: `lettuce`, `countertop`; attributes: `cool`, `sliced` (both on lettuce)
    - "examine bowl with desklamp" → entities: `bowl`, `desklamp`; attributes: (none)

    For each entity noun: search trajectory and grounding observations for a matching instance ID (case-insensitive, rough singular/plural matching). Record OBSERVED status with instance id and location, or NOT_OBSERVED.

    For each attribute modifier: check whether the entity it modifies has that attribute in `objects: attributes=[...]` of the FROM TRAJECTORY section. Record OBSERVED on `<instance>` (if at least one matching entity instance has the attribute) or NOT_OBSERVED.

    This is mechanical string presence — do not infer or guess.

5. **Extract the receptacle list from the initial room overview** — typically the first user message, of the form "you see a bed 1, a desk 2, a desk 1, …". Compute the set of receptacles in that overview list that do NOT appear in the `receptacles:` block of `=== FROM TRAJECTORY ===`. These are the receptacles never visited or never had their contents observed. List them as unvisited.

6. Call `submit_state` once with a single string holding the report in the format below.

## Output format

The report must be a single string with four sections, in this exact order:

```
=== FROM TRAJECTORY ===
agent_self:
  location: <last confirmed location | unknown>  [turn N]
  carrying: <object id | nothing | unknown>  [turn N]
objects:
  - <object id>: last_seen_at=<receptacle id | unknown>, attributes=[<clean|hot|cool|sliced|...> | none], evidence=[turn N(, turn M)]
  - ...
receptacles:
  - <receptacle id>: state=<open|closed|unknown>, contents=[<object id, ...> | not_observed], evidence=[turn N]
  - ...

=== FROM GROUNDING OBSERVATIONS ===
- env_read(<action>) → <verbatim observation excerpt>
- ...

=== TASK VOCABULARY ===
- <task noun>: OBSERVED as <instance id> at <location> | NOT_OBSERVED
- ...

=== UNVISITED RECEPTACLES ===
- <receptacle id>
- ...
```

Conventions:
- Cite trajectory turn numbers in `[turn N]`. If a fact is supported by multiple turns, list each.
- For grounding observations, quote the relevant excerpt of the env_read output verbatim; do not paraphrase.
- Omit the `=== FROM GROUNDING OBSERVATIONS ===` section entirely if you made no env_read calls in this round. Do not write "(none)".
- If a relevant slot has no observation in either source, write `unknown` (and only that). Do not guess, do not fall back to common-sense priors.
- Include in the `objects` and `receptacles` lists only entities the trajectory or your env_read calls have actually surfaced. Do not enumerate hypothetical instances.
- `=== TASK VOCABULARY ===` lists every entity-class noun AND every attribute/state modifier mentioned in the task instruction. Two entry types:
    - **Entity entries** (one per noun): `OBSERVED as <instance> at <loc>` (when at least one matching instance has appeared in observations) or `NOT_OBSERVED`.
    - **Attribute entries** (one per modifier): `OBSERVED on <instance>` (when at least one matching entity instance shows that attribute in the `objects:` block) or `NOT_OBSERVED on any instance`.

    Do not append any explanation or hint. If the task names multiple instances of the same class (e.g., "desk" while the room has `desk 1` and `desk 2`), list each. Tag attribute entries clearly so readers don't confuse them with entity entries — prefix with the attribute word followed by `(attribute on <noun>):`.
- `=== UNVISITED RECEPTACLES ===` is a bare list of receptacle ids drawn from the initial room overview that are absent from the `receptacles:` block above. Write `(none)` if all room receptacles have been observed. Do not annotate, sort, or comment on the list.

## Strict prohibitions

The report must not contain any of the following. If a slot would require any of these to fill, leave it as `unknown`.

- No decision tokens (`APPROVE`, `REJECT`, `OK`, `BLOCK`, `WARN`, etc.).
- No `reason`, `note`, `comment`, or free-form prose section. The schema above is exhaustive.
- No advice, suggestion, or imperative directed at the main agent. No "should", "next", "consider", "recommend", "instead", "before doing X".
- No characterization of the pending action ("the agent intends to ...", "the action would ..."). The pending action is not a topic in the report.
- No inference from common sense or world knowledge ("apples are usually in the fridge", "knives are typically in drawers"). Only direct observation counts.
- No hedged or modal language: no "possibly", "likely", "appears to", "should be", "expected", "probably". Each fact is either observed or absent.
- No reference to past State Reporter rounds beyond the trajectory data itself.
- No interpretation of `=== TASK VOCABULARY ===` results. Do NOT write things like "agent should look elsewhere because pencil is NOT_OBSERVED" or "the apple needed for the task is here". Status entries are bare facts; the agent reads them and decides on its own.
- No annotation of `=== UNVISITED RECEPTACLES ===`. The list is just receptacle ids — no priority labels, no "most likely contains X" suggestions, no commentary.

## Examples

### Example 1 — trajectory already covers everything

Task: "Put a clean apple on the desk."

Room overview (from initial user message): "you see a desk 1, a fridge 1, a sinkbasin 1, a countertop 1, a countertop 2."

Trajectory (excerpted):
- turn 4: `inventory` → "You are carrying: an apple 2."
- turn 6: agent action `clean apple 2 with sinkbasin 1` → "You clean the apple 2 using the sinkbasin 1."
- turn 8: agent action `go to desk 1` → "You arrive at desk 1. On the desk 1, you see a pen 3, a keychain 2."

Pending action: `put apple 2 in/on desk 1`

Task nouns: `apple`, `desk`. Room receptacles: desk 1, fridge 1, sinkbasin 1, countertop 1, countertop 2. Visited (in trajectory): desk 1, sinkbasin 1.

```
=== FROM TRAJECTORY ===
agent_self:
  location: desk 1  [turn 8]
  carrying: apple 2  [turn 4]
objects:
  - apple 2: last_seen_at=in_inventory, attributes=[clean], evidence=[turn 4, turn 6]
  - pen 3: last_seen_at=desk 1, attributes=[none], evidence=[turn 8]
  - keychain 2: last_seen_at=desk 1, attributes=[none], evidence=[turn 8]
receptacles:
  - desk 1: state=open, contents=[pen 3, keychain 2], evidence=[turn 8]
  - sinkbasin 1: state=open, contents=not_observed, evidence=[turn 6]

=== TASK VOCABULARY ===
- apple: OBSERVED as apple 2 at in_inventory
- clean (attribute on apple): OBSERVED on apple 2
- desk: OBSERVED as desk 1 at desk 1

=== UNVISITED RECEPTACLES ===
- fridge 1
- countertop 1
- countertop 2
```

(No `=== FROM GROUNDING OBSERVATIONS ===` section because no env_read was called.)

### Example 2 — trajectory carry state is stale, grounding refreshes via inventory

Task: "Cool a pan and put it on the countertop."

Room overview: "you see a stoveburner 1, a countertop 1, a countertop 2, a fridge 1, a sinkbasin 1, a microwave 1."

Trajectory (excerpted):
- turn 2: `inventory` → "You are not carrying anything."
- turn 5: agent action `take pan 1 from stoveburner 1` → "You pick up the pan 1 from the stoveburner 1."
- turn 7: agent action `take saltshaker 1 from countertop 2` → "Nothing happens." [environment rejected: already carrying]
- turn 8: agent action `go to fridge 1` → "You arrive at fridge 1. The fridge 1 is closed."

Pending action: `cool pan 1 with fridge 1`

Task nouns: `pan`, `countertop`. Visited: stoveburner 1, countertop 2, fridge 1.

```
ASSISTANT (env_read call):
env_read({"action": "inventory"})

TOOL:
"You are carrying: a pan 1."

ASSISTANT (submit_state call):
submit_state({"report": "...as below..."})
=== FROM TRAJECTORY ===
agent_self:
  location: fridge 1  [turn 8]
  carrying: unknown
objects:
  - pan 1: last_seen_at=stoveburner 1, attributes=[none], evidence=[turn 5]
  - saltshaker 1: last_seen_at=countertop 2, attributes=[none], evidence=[turn 7]
receptacles:
  - stoveburner 1: state=open, contents=[pan 1 (taken at turn 5)], evidence=[turn 5]
  - countertop 2: state=open, contents=[saltshaker 1], evidence=[turn 7]
  - fridge 1: state=closed, contents=not_observed, evidence=[turn 8]

=== FROM GROUNDING OBSERVATIONS ===
- env_read(inventory) → "You are carrying: a pan 1."

=== TASK VOCABULARY ===
- pan: OBSERVED as pan 1 at in_inventory (env_read inventory)
- cool (attribute on pan): NOT_OBSERVED on any instance
- countertop: OBSERVED as countertop 2 at countertop 2

=== UNVISITED RECEPTACLES ===
- countertop 1
- sinkbasin 1
- microwave 1
```

### Example 3 — target receptacle contents not yet observed; task noun never seen

Task: "Look at pencil under the desklamp."

Room overview: "you see a desk 1, a desk 2, a drawer 1, a drawer 2, a shelf 1, a shelf 2, a bed 1, a garbagecan 1."

Trajectory (excerpted):
- turn 1: agent action `go to desk 1` → "On the desk 1, you see a desklamp 1, a pen 2, a laptop 1."
- turn 2: agent action `examine desk 1` → "On the desk 1, you see a desklamp 1, a pen 2, a laptop 1."
- turn 3: agent action `use desklamp 1` → "You turn on the desklamp 1."
- turn 5: agent action `take pen 2 from desk 1` → "You pick up the pen 2 from the desk 1."

Pending action: `examine desk 1`

Task nouns: `pencil`, `desklamp`. The trajectory never shows any `pencil` instance — only `pen 2`, which is a distinct entity class. Visited: desk 1.

```
=== FROM TRAJECTORY ===
agent_self:
  location: desk 1  [turn 1, turn 2, turn 3]
  carrying: pen 2  [turn 5]
objects:
  - desklamp 1: last_seen_at=desk 1, attributes=[none], evidence=[turn 1, turn 3]
  - pen 2: last_seen_at=in_inventory, attributes=[none], evidence=[turn 1, turn 5]
  - laptop 1: last_seen_at=desk 1, attributes=[none], evidence=[turn 1]
receptacles:
  - desk 1: state=open, contents=[desklamp 1, pen 2 (taken at turn 5), laptop 1], evidence=[turn 1, turn 2]

=== TASK VOCABULARY ===
- pencil: NOT_OBSERVED
- desklamp: OBSERVED as desklamp 1 at desk 1

=== UNVISITED RECEPTACLES ===
- desk 2
- drawer 1
- drawer 2
- shelf 1
- shelf 2
- bed 1
- garbagecan 1
```
\end{Verbatim}
\end{tcolorbox}

A separate policy-level grounding prompt $\pi_{\mathrm{gp}}$
(\S\ref{sec:grounding-prompt}) is appended to each main agent's
system prompt to bias it toward inspecting environmental content
before acting. The exact text differs between benchmarks, scaled
to each environment's interaction surface.

\begin{tcolorbox}[aibox, breakable, title=AppWorld policy-level grounding prompt, label=box:grounding-prompt-policy-appworld]
\textbf{(10)} If the content or format of returned information is
uncertain, inspect the actual raw content first (e.g., via
\texttt{read}/\texttt{show} APIs) and do not guess structure or
write parsing logic based on assumptions.

\medskip
\textbf{(11)} If the content of returned information is not as
expected, re-examine the API specification, your request
parameters, and any assumptions you made about the response
structure. Adjust your approach accordingly and verify the result
is correct before proceeding to the next step.

\medskip
\textbf{(12)} Before executing the task, you MUST call
\texttt{show\_api\_descriptions(app\_name="")} to get the complete
API list for ALL apps at once. Do NOT selectively check only apps
you think are relevant---you must examine every app's APIs
exhaustively.
\end{tcolorbox}

\begin{tcolorbox}[aibox, breakable, title=ALFWorld policy-level grounding prompt, label=box:grounding-prompt-policy-alfworld]
Before you start to tackle the problem, explore the environment
first.
\end{tcolorbox}

\end{document}